\documentclass[10pt,twocolumn,letterpaper]{article}

\usepackage[pagenumbers]{cvpr} 

\usepackage[dvipsnames]{xcolor}

\definecolor{cvprblue}{rgb}{0.21,0.49,0.74}
\usepackage[pagebackref,breaklinks,colorlinks,citecolor=cvprblue]{hyperref}
\usepackage{multirow}
\usepackage{multicol}

\def\paperID{1577} 
\def\confName{CVPR}
\def\confYear{2024}

\title{DiffCTE: Consistent Style Text Editing with Diffusion Model}

\author{Xiangxiang Wang, Huang Chen, Bing Liu, Haoyu Cao, Xin Li, Yinsong Liu, Deqiang Jiang, Xing Sun\\
Tencent YouTu Lab\\
}

\begin{document}
\maketitle
\begin{abstract}

Text image editing is a critical task in computer vision and graphics research, with diverse applications such as advertising design and social media content creation.
Generating high-quality text images with readability and consistency using Diffusion Models (DFMs) poses a challenge due to their inadequate handling of text styles and glyphs.
In this paper, we present DiffCTE (Consistent Style Text Editing with Diffusion Model), a novel approach that enhances the generation of readable and stylistically consistent text images by providing DFMs with improved style condition guidance.
Leveraging a vision language model (VLM) consisting of a Vision Encoder, Style Abstractor, and Text Decoder, the Style Abstractor effectively extracts style condition and represents style information across languages, encompassing aspects such as glyph, font and color.
By integrating the style condition into DFMs, DiffCTE enables visually coherent and stylistically consistent text generation.
Specifically, the generated text achieves style transfer by maintaining consistency with the reference style.
Our comprehensive experiments demonstrate the superior performance of DiffCTE in comparison to existing methods, setting new directions for future research in text image editing with DFMs.
    
\end{abstract}    
\section{Introduction}
\label{sec:intro}

Text image editing has become a crucial task in computer vision and graphics due to its broad applicability in domains such as advertising design, social media content creation, and information visualization~\cite{cao2023mobile,huang2022gentext,shi2022aprnet,wu2019editing,kwon2022clipstyler,liu2020erasenet}. 
The primary objective is to modify or add specific text within an image while preserving visually coherent and natural appearances, enabling personalized and diverse information delivery, and consequently enhancing visual appeal and communication effectiveness.

Recent advancements in image generation techniques have witnessed a transition from Generative Adversarial Networks (GANs)~\cite{zhan2019spatial,luo2021separating,wu2019editing,roy2020stefann,qu2023exploring,su2023scene,liu2020erasenet,kang2021content,luo2021separating,yu2021mask,lee2021rewritenet} to Diffusion Models (DFMs)~\cite{rombach2022high,ji2023improving,DiffUTE,wu2022unifying,wang2023letter,chen2023textdiffuser,hertz2022prompt,ma2023glyphdraw,tuo2023anytext,li2023blip} owing to DFMs' ability to produce high-quality and more realistic natural scene images. 
Nevertheless, DFMs face challenges in generating quality text and document images that exhibit style consistency. 
This limitation arises from their restricted understanding of text styles and glyphs, constraining their versatility and impact across various text-involved applications.
\begin{figure}
\centering
    \includegraphics[width=0.5\textwidth,trim=250 120 160 50, clip]{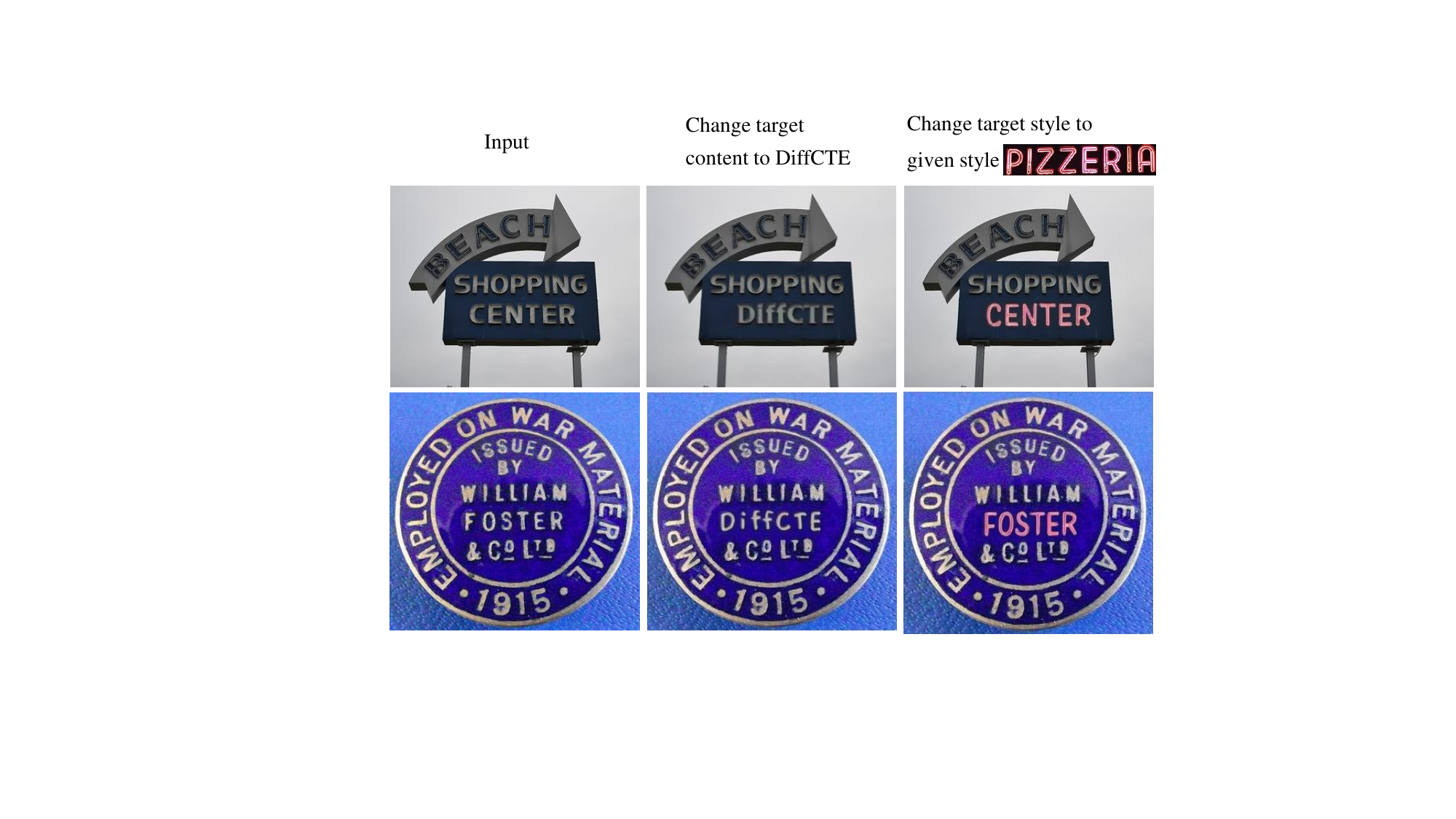}
    \caption{Visualization of DiffCTE capabilities. DiffCTE excels in: (a) Scene text editing: Modifying target content while maintaining the original image style, resulting in consistent and natural text editing. (b) Text style transfer: Adjusting target style based on a reference image, effectively transferring the style with a consistent appearance.}
    \label{fig:intro_vis}
    \vspace{-0.50cm}
\end{figure}

To address these challenges, we propose DiffCTE (\textbf{C}onsistent Style \textbf{T}ext \textbf{E}diting with \textbf{Diff}usion Model), a novel approach designed to provide better style condition guidance for DFMs, focusing on style consistency. 
DiffCTE employs a pre-trained VLM~\cite{kim2022ocr}, which exhibits exceptional performance in the document understanding domain. The VLM is composed of a Vision Encoder, Style Abstractor, and Text Decoder, which collectively work to enhance the readability and style coherence of generated images.
As illustrated in the Figure~\ref{fig:intro_vis}, DiffCTE achieves consistency in style with either the original image or the reference image.

The \textbf{Style Abstractor}, a key component of DiffCTE, is a pre-trained style encoder proficient in converting specified text styles into fixed-dimension vectors. 
It accounts for various aspects such as font, weight, slant, and color, enabling the DFM to generate consistent text with the desired style. 
Moreover, the Style Abstractor addresses the fine-grained nature of glyphs in languages, thus improving the model's understanding of glyph representations and the reconstruction of visually coherent text images.

To further enhance the performance of the Style Abstractor, it undergoes training on multiple tasks, including editing region prediction, editing text recognition, and style extraction.
This training process allows the Style Abstractor to accurately and adaptively extract style conditions from diverse sources, applying them effectively to the DFM.

As shown in Figure~\ref{fig:intro_vis}, what sets DiffCTE apart from existing methods is its ability to generate text with style consistency derived from original image or other reference images, resulting in more realistic and identifiable text images that cater to a wide range of applications.
In summary, our main contributions are:

\begin{itemize}
\item The proposal of DiffCTE, a novel text image editing approach that provides better style condition guidance for DFMs, focusing on style consistency and thus enhancing readability and style coherence in generated images.
\item The introduction of the Style Abstractor, adept at extracting and representing text styles across various languages while considering fine-grained glyph representations.
\item A comprehensive experimental evaluation demonstrating the effectiveness of DiffCTE and its superiority compared to existing methods in various text-involved applications, particularly in addressing the issue of style consistency, including style transfer.
\end{itemize}

\section{Related Work}

Scene text editing, a research area focusing on altering the original text content within an image while maintaining its style, has drawn considerable attention in recent years. 
The primary approaches for scene text editing comprise Generative Adversarial Networks (GANs) and Diffusion models.

\subsection{GAN-based Methods}
GAN-based methods for scene text editing~\cite{wu2019editing,zhan2019spatial,roy2020stefann,yang2020swaptext,shimoda2021rendering,kang2021content,luo2021separating,yu2021mask,lee2021rewritenet,huang2022gentext,kong2022look} have gained widespread popularity, which view scene text editing as a style transfer issue. 
The approaches necessitate the use of images as desired style templates, typically a cropped image portion containing target text. 
The main objective is rendering the desired text to match the reference style and background. 
SF-GAN~\cite{zhan2019spatial} employs GANs to enhance synthesis realism in geometry and appearance spaces during style transfer. 
SRNet~\cite{wu2019editing}, which employs a modular and joint-trainable sub-net structure that divides the task into separate sub-tasks (e.g., extracting the image background and text spatial alignment) for an end-to-end pipeline. 
STEFANN~\cite{roy2020stefann} operates at the character level, utilizing a font-adaptive neural network for font structure extraction and a color-preserving model for color extraction, resulting in more flexible text adjustments. 
MOSTEL~\cite{qu2023exploring}, which incorporates stroke-level information, guided prediction maps, and explicit background filtering to achieve more accurate style transfer and scene text editing. 
QuadNet~\cite{su2023scene} combines background inpainting, style encoding, content encoding, and fusion generation, enabling various text style modifications, such as rotation angle, font type, and text color.
Despite achieving progress, GAN-based methods experience constraints in handling arbitrary font styles, sizes, and colors, often yielding in unnatural and inconsistent outcomes between the edited and surrounding areas.

\subsection{Diffusion Model-based Methods} 
Diffusion models have demonstrated exceptional capabilities in generating natural and realistic scenes, making remarkable advancements in the field of image editing~\cite{balaji2022ediffi,brack2023sega,ho2020denoising,li2023guiding,nichol2021glide,nichol2021improved,ryu2022pyramidal,wu2022unifying,rombach2022high,wu2023uncovering,xie2023smartbrush,brooks2023instructpix2pix,DiffUTE,ji2023improving,ma2023glyphdraw,tuo2023anytext}, 
particularly Stable Diffusion (SD) models~\cite{rombach2022high}, which attain a near-optimal trade-off between complexity reduction and detail preservation by applying diffusion models in the latent space of pre-trained auto-encoders. 
Recently, specialized diffusion models have emerged focusing on text scene editing. These models provide more text-specific constraints for enhanced detail generation.
DiffSTE~\cite{ji2023improving} builds upon pretrained diffusion models featuring a dual encoder design for improved text legibility and style control, using separate encoders for character information and visual guidance. 
DiffUTE~\cite{DiffUTE} generates high-quality text through fine-grained control of glyph and position information, combining character embeddings and powerful pre-trained auto-encoders to create realistic text editing results.
But, it underperforms when dealing with complex glyph.
However, it has been observed that diffusion models exhibit suboptimal performance in generating visual text at specific locations, often resulting in the creation of incongruous text styles.
There is a significant dearth of in-depth research dedicated to augmenting the proficiency of diffusion models in generating consistent text styles within images.
The primary focus of our proposed method, DiffCTE, is to address the limitations faced by GAN-based and diffusion model-based scene text editing methods in generating consistent text styles within images. 
\section{Methodology}

\begin{figure*}
  \centering
  \includegraphics[width=0.96\textwidth,trim=0 0 0 0, clip]{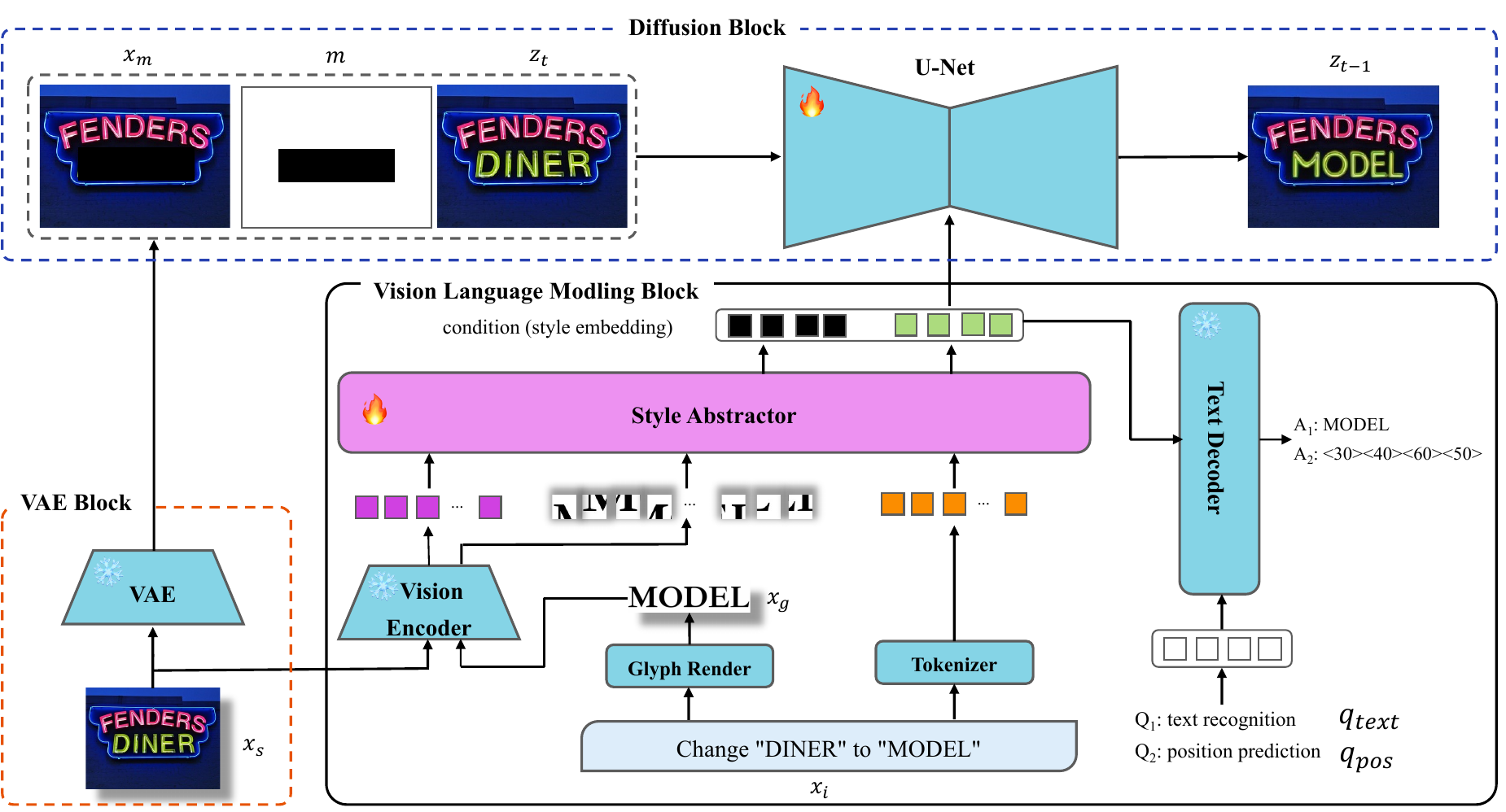}
  \vspace{-0.2cm}
  \caption{The DiffCTE framework comprises three main components: VAE, U-Net, and VLM. The process begins with the VAE block encoding the input image $x_s$ into a latent space $z_t$. Next, the Diffusion block reconstructs the expected image utilizing the style embedding extracted by the Style Abstractor, as well as the image latent vector $z_t$, mask $m$, and masked latent image vector $x_m$. Concurrently, the VLM block enhances the style embedding by integrating the input image $x_s$ and instruction text $x_i$. The Glyph Render, processing the instruction text $x_i$, generates a glyph image $x_g$, which is subsequently divided into patches. The Vision Encoder then transforms the input image $x_s$ into visual embeddings. By combining these visual embeddings, image patches, and text tokens, the Style Abstractor outputs a refined style embedding, which is employed by the Text Decoder for accurate localization and recognition.}
  \label{fig:framework}
  \vspace{-0.40cm}
\end{figure*}

In this section, we introduce DiffCTE, a novel text editing framework that combines a VLM and a diffusion model. 
The VLM has demonstrated competitive performance in tasks such as document information extraction and text spotting~\cite{kim2022ocr}.
The complete framework is depicted in Figure~\ref{fig:framework}.
We build upon the foundation of Stable Diffusion (SD)~\cite{rombach2022high} and incorporate additional style condition constraints to enhance its text editing capabilities. 
The refined model architecture and training process are detailed in the subsequent subsections.
The training process comprises both supervised training using synthetic data and self-supervised training using realistic data.

\subsection{Diffusion Model with Condition}
SD models employ a variational auto-encoder (VAE) to improve computational efficiency by performing the diffusion process within a low-dimensional latent space.
Given an input image $\mathbf{x} \in \mathbb{R}^{H \times W \times 3}$, the VAE encoder $\mathcal{E}_v$ transforms it into a latent representation $\mathbf{z} \in \mathbb{R}^{h \times w \times c}$.
The diffusion process is performed in the latent space, utilizing a conditional U-Net denoiser $\epsilon_\theta(\mathbf{z_t}, t, \mathbf{y})$, which operates with a noisy latent $\mathbf{z_t}$, a generation condition input $\mathbf{y}$, and the current time step $t$. Various modalities, such as natural language or segmentation maps, can serve as condition information $\mathbf{y}$.

To pre-process $\mathbf{y}$ from different modalities, a domain-specific encoder, $\tau_\theta$, projects it into an intermediate representation $\tau_\theta(\mathbf{y}) \in \mathbb{R}^{M \times d_\tau}$.
This representation is then mapped to the intermediate layers of the U-Net through a cross-attention mechanism.
In text-to-image generation scenarios, a pre-trained CLIP text encoder $\tau_\theta$ generates condition $C=\tau_\theta(\mathbf{y})$ by encoding text prompts $\mathbf{y}$.

The overall training objective of SD is defined as:
\vspace{-0.10cm}
\begin{equation}
    \mathcal{L}_{sd} = \mathbb{E}_{\mathcal{E}(\mathbf{x}),\mathbf{y},\epsilon\sim \mathcal{N}(0,1),t} \left[
    \lVert
    \epsilon-\epsilon_\theta(\mathbf{z_t},t,\tau_\theta(\mathbf{y}))
    \rVert^2_2\right].
    \label{eq:obj}
\end{equation}

Our DiffCTE framework builds upon this foundation and focuses on optimizing encoders $\tau_\theta$ and $\epsilon_\theta$.
By integrating the additional style condition constraints, our method aims to address the limitations faced by existing diffusion models in generating consistent text styles within images.


\subsection{Style Abstractor}

Our objective is to enable the diffusion model to accurately generate the desired editing text with appropriate glyphs while maintaining style consistency and improving the readability of the editing region. 
To achieve this, we incorporate a VLM in DiffCTE, consisting of a vision encoder \(\mathrm{ENC}_{vis}\), a style abstractor \(\mathrm{ABS}_{style}\), and a text decoder \(\mathrm{DEC}_{text}\), as depicted in Figure \ref{fig:framework}. 
The VLM is proficient in extracting style information and utilizing text queries to recognize textual content and predict relevant text locations within an image.

The \textbf{vision encoder} \(\mathrm{ENC}_{vis}\) leverages a vision transformer backbone to extract image features \(\mathbf{f_s}\).
The style abstractor \(\mathrm{ABS}_{style}\) bridges the gap between the vision encoder and the text decoder, and the style abstractor is composed of multiple transformer layers.
Within these layers, the glyph image encodings \(\mathbf{f_g}\) and the text instructions \(\mathbf{f_i}\) interact individually with the image features \(\mathbf{f_s}\) through cross-attention layers, enabling the style abstractor to generate embeddings that capture and represent style information.

The \textbf{text decoder} \(\mathrm{DEC}_{text}\), a transformer decoder, encodes input queries and attends to the style embeddings.
It generates text in an auto-regressive manner by attending to the query embeddings, the style embeddings, and previously generated tokens.
The mathematical form of the text decoder is as follows:

\vspace{-0.28cm}
\begin{equation}
  \mathbf{h_t} = \mathrm{Decoder}(\mathbf{h_{t-1}}, \mathbf{f_{g}}, \mathbf{q}, \mathbf{\tau_{t-1}})
  \label{eq:text_decoder_gen}
\end{equation}

Here, \(\mathbf{h_t}\) represents the hidden state of the decoder at time step \(t\), \(\mathbf{f_g}\) represents the style embeddings, \(\mathbf{q}\) represents the query embedding, and \(\mathbf{\tau_{t-1}}\) represents the previous token embedding.
Depending on different queries, the text decoder can perform position prediction and text recognition tasks for editing regions or encoding glyph images.
As shown in Figure~\ref{fig:framework}, when the query is ``text recognition'', the text decoder will conduct text recognition on \(\mathbf{x_g}\) or the editing region within the provided inputs \(\mathbf{x_s}\).
Conversely, when the query is ``position prediction'', the text decoder will predict the position of the editing region.
In Figure~\ref{fig:framework}, the bounding box $<x_1><y_1><x_2><y_2>$ represent the predicted position tokens of editing region.

This approach resembles the cross-attention mechanism in Stable Diffusion, with a modification~\cite{rombach2022high,DiffUTE}. 
Instead of the original input latent vector \(\mathbf{{z}_t}\), we replace it with the concatenation of the latent image vector \(\mathbf{{z}_t}\), the masked image latent vector \(\mathbf{{x}_m}\), and the text mask \(\mathbf{m}\), yielding a more comprehensive representation of the input information.
Furthermore, the style abstractor extracts embeddings that represent the style by utilizing the original image features, glyph image features, and text instructions from the input, which are referred to as the style condition.
These embeddings capture style characteristics and facilitate fine-grained diffusion control during the text editing process.
Through this integration, we significantly enhance the diffusion model's ability to generate text edits with improved style consistency and superior generative performance.

\subsection{Supervised Training Strategy}

To improve the extraction capabilities of the Style Abstractor in the DiffCTE framework, we leverage a synthetic dataset and a tailored VLM for training purposes.
The synthetic dataset is generated by randomly selecting numbers, English words, or Chinese characters and creating paired image data with consistent font styles, colors, and overall visual appearance, which is similar to the data synthesis method in DiffSTE~\cite{ji2023improving}.
In order to consider style transfer from a reference image, it is necessary to construct paired images that contain the same textual content but exhibit variations in text styles, and one of the paired images aligns with the style of the reference image.

The DiffCTE is trained using this synthetic dataset, consisting of tuples \(\{(\mathbf{x_s},\mathbf{x_g},\mathbf{x_i},\mathbf{q_{pos}},\mathbf{q_{text}}),\mathbf{x_t}\}\), which encompass input images, glyph images, text instructions, text recognition queries, position prediction queries for the editing region, and target images.
When \(\mathbf{x_g}\) is text image by generating with a uniform font style, \(\mathbf{x_s}\) and \(\mathbf{x_t}\) are paired images, where both share the same style but differ in their textual content.
Conversely, when \(\mathbf{x_g}\) is text image that represents reference style, \(\mathbf{x_s}\) and \(\mathbf{x_t}\) are paired images, where both share the same textual content but differ in their text style.
This training data allows us to refine the Style Abstractor's extraction effectiveness within the VLM significantly.
Specifically, it allows us to extract style information from either \(\mathbf{x_s}\) or \(\mathbf{x_g}\), depending on the given type of \(\mathbf{x_g}\).

By adopting a training strategy that only trains the parameters of the Style Abstractor and the U-Net, and the mean squared error (MSE) loss between the ground truth image and the generated image is retained, which ensures that the output \(\mathbf{f_g}\) of the Style Abstractor accurately represents style information.
Furthermore, the Style Abstractor is capable of generating corresponding style conditions when different inputs \(\mathbf{x_g}\) are provided.

In the VLM, we aim to enhance the positional accuracy of the editing region and the text recognition within it while preserving complete style information in the style embeddings.
To achieve this, we introduce separate cross-entropy losses for different queries in the text decoder.
These losses account for the positional coordinates and textual content within the editing region, leading to improved prediction precision in the model.

Through this enhanced DiffCTE training process, we bolster the capabilities of DiffCTE framework, ultimately resulting in more visually appealing and contextually coherent text editing within images.

\begin{table*}
  \centering
  \renewcommand{\arraystretch}{1.00}
  \setlength{\tabcolsep}{3.5mm}
  \begin{tabular}{@{}c|cc|cc|cc|cc|cc@{}}
    \toprule
    \multirow{2}*{Model} & \multicolumn{2}{|c}{ArT} & \multicolumn{2}{|c}{COCOText} & \multicolumn{2}{|c}{TextOCR} & \multicolumn{2}{|c}{ICDAR13} & \multicolumn{2}{|c}{Average} \\
    ~ & OCR & Cor & OCR & Cor & OCR & Cor & OCR & Cor & OCR & Cor \\
    \midrule
    SRNet~\cite{wu2019editing}      & 30.63 & 42 & 27.84 & 73 & 32.58 & 42 & 32.55 & 46 & 30.90 & 50.75 \\
    MOSTEL~\cite{qu2023exploring}   & 60.72 & 70 & 27.67 & 74 & 47.33 & 56 & 50.99 & 58 & 46.68 & 64.50 \\
    SD1~\cite{rombach2022high}      &  5.67 &  6 & 11.97 & 10 &  9.07 &  4 &  5.13 &  6 &  7.96 &  6.50 \\
    SD2~\cite{rombach2022high}      &  8.19 & 17 & 19.73 & 14 & 10.63 & 14 &  9.59 &  9 & 12.04 & 13.50 \\
    SD1-FT~\cite{rombach2022high}   & 31.04 & 32 & 52.97 & 60 & 49.23 & 44 & 25.93 & 30 & 39.79 & 41.50 \\
    SD2-FT~\cite{rombach2022high}   & 48.72 & 45 & 56.72 & 50 & 61.47 & 58 & 44.13 & 47 & 52.76 & 50.00 \\
    DiffSTE~\cite{ji2023improving}  & 84.10 & 87 & 83.78 & 87 & 85.59 & 82 & 82.07 & 85 & 83.89 & 85.25 \\
    \midrule
    DiffCTE  & \textbf{86.29} & \textbf{89} & \textbf{84.74} & \textbf{88} & \textbf{88.17} & \textbf{88} & \textbf{83.32} & \textbf{86} & \textbf{85.63} & \textbf{87.75} \\
    \bottomrule
  \end{tabular}
  \caption{Quantitative evaluation results across four datasets. Text correctness rate is assessed using both the OCR model (OCR) and human evaluation (Cor). Bold text indicates superior results.}
  \label{tab:quat_res}
\end{table*}

\begin{table*}
  \centering
  \renewcommand{\arraystretch}{1.00}
  \setlength{\tabcolsep}{3.5mm}
  \begin{tabular}{@{}c|cc|cc|cc|cc|cc@{}}
    \toprule
    \multirow{2}*{Model} & \multicolumn{2}{|c}{ArT} & \multicolumn{2}{|c}{COCOText} & \multicolumn{2}{|c}{TextOCR} & \multicolumn{2}{|c}{ICDAR13} & \multicolumn{2}{|c}{Average} \\
    ~ & OCR & Cor & OCR & Cor & OCR & Cor & OCR & Cor & OCR & Cor \\
    \midrule
    w/o Pos. & 87.35 & 90 & 85.03 & 89 & 88.29 & 88 & 82.89 & 85 & 85.89 & 88.00\\
    w/o Sty.    & 51.32 & 47 & 46.73 & 44 & 63.11 & 59 & 46.57 & 45 & 51.93 & 48.75 \\
    \midrule
    DiffCTE      & 86.29 & 89 & 84.74 & 88 & 88.17 & 88 & 83.32 & 86 & 85.63 & 87.75 \\
    \bottomrule
  \end{tabular}
  \caption{Ablation study results across four datasets, examining the impact of position prediction (Pos.) and style condition (Sty.) on the model's performance.}
  \label{tab:ablation_res}
  \vspace{-0.3cm}
\end{table*}

\subsection{Self-supervised Training Strategy}
\label{subsec:self-super}

Collecting and annotating large-scale datasets with realistic paired data for text image editing is extremely challenging and impractical due to the intensive resources, labor, and costs involved.
To address this issue, we propose a self-supervised training approach for the diffusion model inspired by DiffUTE~\cite{DiffUTE}.

In our strategy, the training data consists of tuples \(\{(\mathbf{x_s}, \mathbf{x_m}, \mathbf{m}, \mathbf{x_g}, \mathbf{x_i}, \mathbf{q_{pos}}), \mathbf{x_s}\}\), containing original image, masked image, text mask, glyph image, text instruction, and query for position prediction. The self-supervised training process is as follows:

\begin{enumerate}
    \item The Style Abstractor processes the glyph image \(\mathbf{x_g}\), which is generated with a uniform font style, along with the instruction \(\mathbf{x_i}\) and the original image \(\mathbf{x_s}\), in order to acquire the style embeddings \(\mathbf{f_g}\).
    \item The text decoder generates the position of the editing region, represented as the text mask \(\mathbf{m}\), using the style embeddings \(\mathbf{f_g}\) and the specific query \(\mathbf{q_{pos}}\) as inputs.
    \item An OCR region or predicted editing region within the image is randomly selected to serve as the text mask.
    \item Create a new latent vector \(\mathbf{z^{\textquotesingle}_t}=\mathrm{Concat}(\mathbf{x_m},\mathbf{m},\mathbf{z_t})\) by concatenating the masked image latent vector \(\mathbf{x_m}\), the text mask \(\mathbf{m}\), and the noisy image latent vector \(\mathbf{z_t}\).
    \item Train DiffCTE to minimize the MSE loss, \(L_{d}\), between the ground truth image and the generated image conditioned on the text instruction, the glyph image, and the masked image.
\end{enumerate}

By adopting this self-supervised training strategy, we can overcome the challenges of collecting and annotating paired data for text image editing.
This approach allows us to effectively train the DiffCTE model without manual annotation, significantly reducing resource requirements and laborious efforts involved in the process.

Based on the training strategy for the Style Abstractor and the self-supervised training approach for the diffusion model, we construct a comprehensive training dataset by merging the synthetic data and realistic data.
This approach involves training the Style Abstractor to improve style extraction using synthetic data and enhancing the diffusion model's text image generation performance with realistic data.
Consequently, we perform end-to-end training of our proposed DiffCTE model with these data, while enhancing the overall performance of our proposed framework.

\section{Experiments}

\begin{figure*}
  \centering
  \includegraphics[width=0.95\textwidth,trim=246 45 250 11, clip]{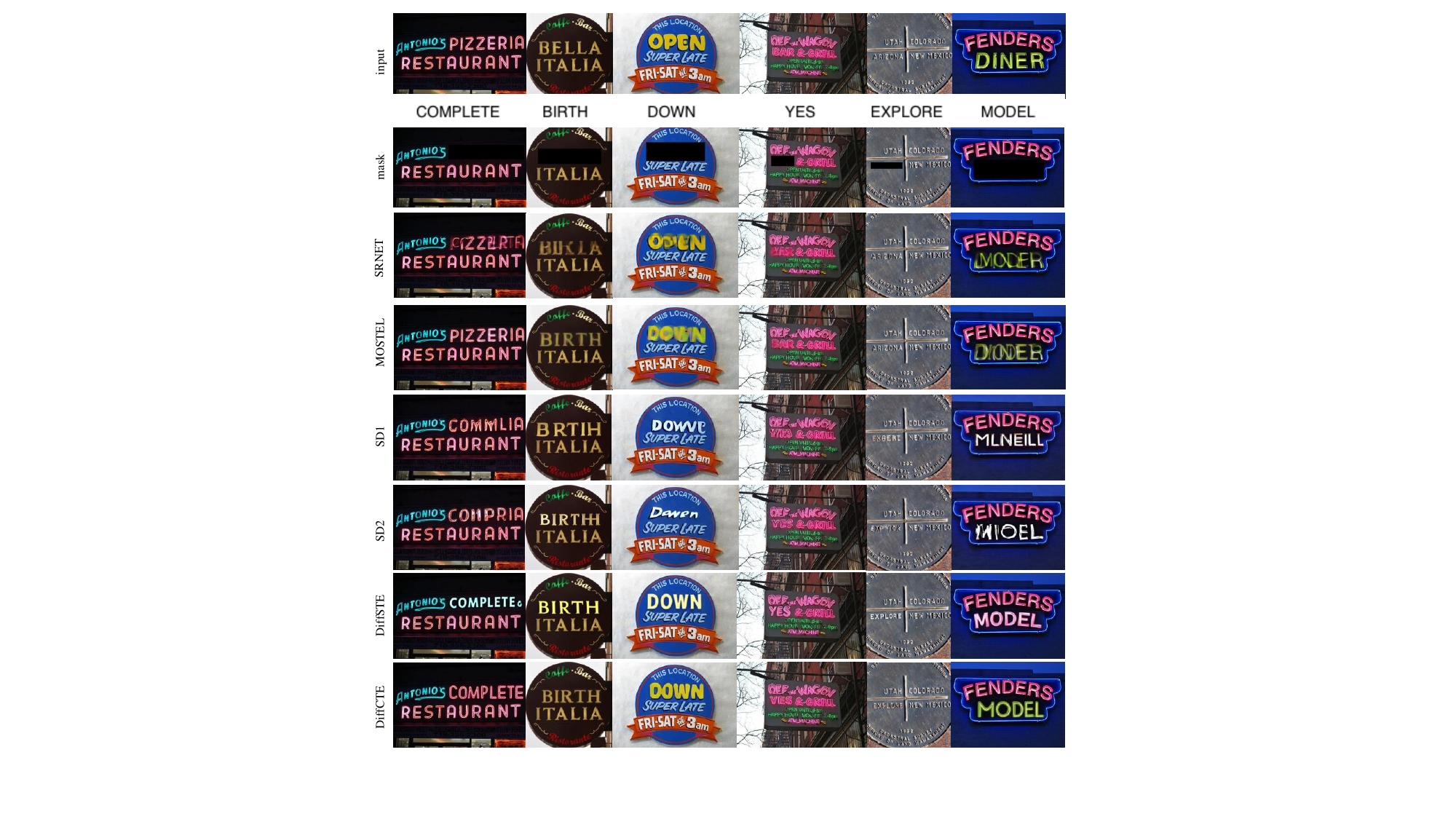}
  \caption{Visualization examples from our text editing experiment. Different rows showcase the text editing results obtained by various methods, with the final row emphasizing the outcome of our proposed approach. Our DiffCTE demonstrates a substantial improvement over other methods, affirming its capability to preserve stylistic consistency in the generated text.}
  \label{fig:vis}
  \vspace{-0.47cm}
\end{figure*}

\begin{figure*}
  \centering
  \includegraphics[width=0.95\textwidth,trim=40 20 40 5, clip]{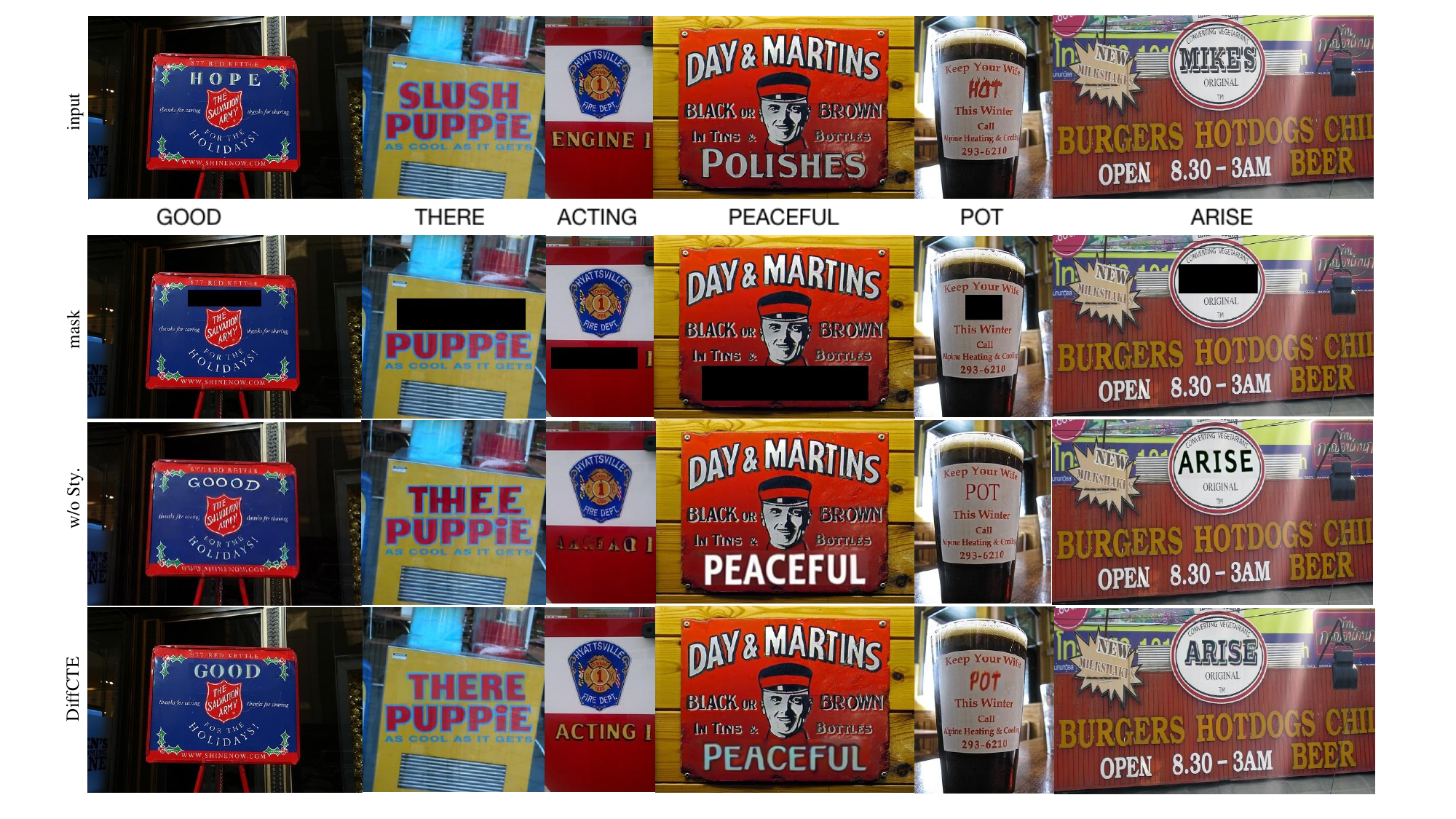}
  \caption{Visualization examples from our text editing experiment without the style condition. Substituting the style embedding from the Style Abstractor with the latent vector obtained from CLIP leads to a decline in the model's capacity to generate accurate text while maintaining style consistency.}
  \label{fig:without_sty}
  \vspace{-0.47cm}
\end{figure*}

\subsection{Datasets for Text Image Generation}

To overcome the challenge of limited large-scale datasets for text image generation, we carefully curated a diverse dataset comprising a substantial number of images. This comprehensive dataset is formed by combining the synthetic dataset with several publicly available text image datasets, including CLDA~\cite{li2021cdla}, XFUND~\cite{xu2022xfund}, CORD~\cite{park2019cord}, Ticket~\cite{guo2019eaten}, PubLayNet~\cite{zhong2019publaynet}, and ICDAR~\cite{zhang2019icdar,nayef2019icdar2019,karatzas2015icdar,huang2019icdar2019,chng2019icdar2019,gomez2017icdar2017}.
To ensure a comprehensive evaluation~\cite{ji2023improving,DiffUTE}, we randomly extracted 1000 samples from individual datasets such as ArT~\cite{chng2019icdar2019}, COCOText~\cite{gomez2017icdar2017}, TextOCR~\cite{singh2021textocr}, and ICDAR13~\cite{karatzas2015icdar} to create distinct test sets for each dataset.

By amalgamating these diverse datasets, we have obtained a wide-ranging training dataset that covers various text image variations. This carefully curated dataset serves as a valuable resource for training and evaluating our text image generation model, bolstering its robustness and generalization capabilities.

\subsection{Implementation Details}
Our proposed DiffCTE model consists of three key components: the VLM, VAE, and U-Net. These components work synergistically to enhance the overall functionality and performance of the model.

To improve the text image reconstruction ability, we conducted a fine-tuning process on the VAE. 
We initialized the VAE using a checkpoint from the stable-diffusion-2 inpainting model\footnote{https://huggingface.co/stabilityai/stable-diffusion-2-inpainting} to enhance its performance in generating accurate and visually appealing text image reconstructions. 
Moreover, we utilize the Style Abstractor from a pre-trained VLM as our style extractor. 
During the pre-training process, the VAE and VLM are trained under optimized settings to ensure effective learning of text image reconstruction and style extraction.
The VAE is trained for 5 epochs with a batch size of 64 and a learning rate of 1e-5.
The VLM is trained for 6 epochs with a batch size of 64 and a learning rate of 3e-5.

Notably, the weights of the VAE, the vision encoder and the text decoder remain frozen during DiffCTE training. 
By maintaining their weights fixed, these pre-trained components bolster DiffCTE's performance while leveraging their learned representations. 
For DiffCTE training, we set the learning rate to 5e-5 and the batch size to 64, optimizing the model's performance based on factors such as computational resources and convergence speed.

The text correctness is employed as a metric to evaluate the accuracy of the generated text.
To measure OCR accuracy, we calculate the exact match rate between the target text and the recognized text generated by a pretrained recognition model.
This quantitative measure provides an assessment of how well the generated text aligns with the ground truth.
In addition, the Cor is quantified through human evaluation.
And human evaluators assess the accuracy of the generated text by comparing it to the ground truth.

By integrating OCR accuracy and human evaluation, our objective is to conduct a comprehensive assessment of the accuracy of the generated text. This methodology enables us to take into account both automated metrics and human judgment, thereby ensuring a robust evaluation of the textual correctness in our research.

Our experiments involve comparative analysis between DiffCTE and other text editing approaches, focusing on scene text editing methods (SRNet~\cite{wu2019editing} and MOSTEL~\cite{qu2023exploring}) and diffusion models (SD1~\cite{rombach2022high}, SD2~\cite{rombach2022high}, DiffSTE~\cite{ji2023improving}). Fine-tuning of SRNet and MOSTEL is performed using synthetic datasets and our own dataset, respectively. Both the SD1 and SD2 models are fine-tuned with instruction tuning (denoted as SD1-FT and SD2-FT). Additionally, we conducted direct comparisons between our models and state-of-the-art text editing approaches utilizing diffusion models, including DiffSTE.

\begin{figure*}
  \centering
  \includegraphics[width=0.85\textwidth,trim=140 115 70 140, clip]{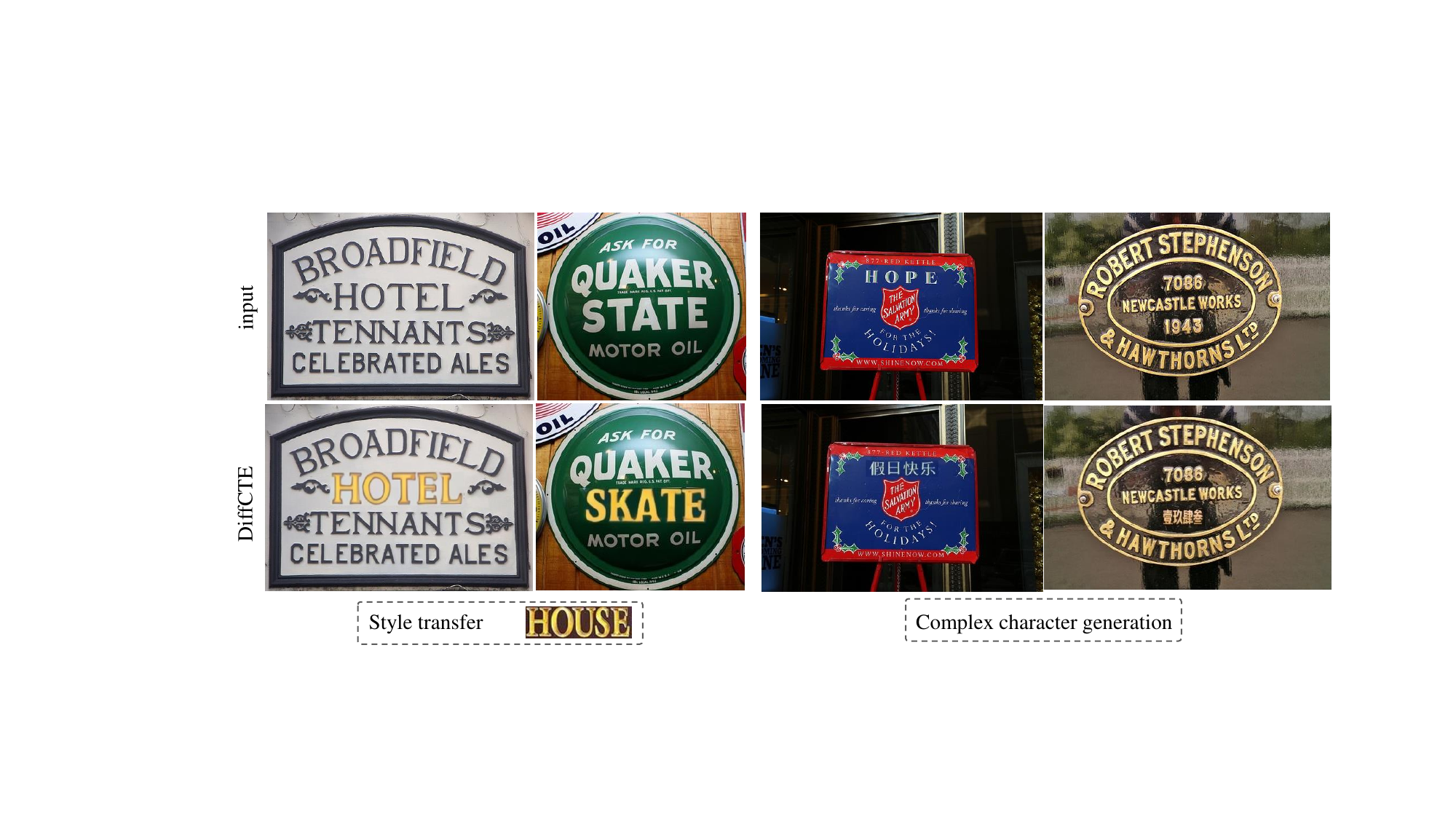}
  \caption{Visualization examples from style transfer tasks and complex character shape generation. The model effectively incorporates the style from the reference image while preserving textual content. Additionally, it generates intricate characters while maintaining the original style, showcasing its versatility and adaptability.}
  \label{fig:ablation_task}
  \vspace{-0.47cm}
\end{figure*}

\subsection{Comparative Evaluation and Results}

The quantitative results of text editing are presented in Table~\ref{tab:quat_res}. 
In comparison with the previous state-of-the-art model, DiffSTE, our proposed DiffCTE shows notable improvements. 
It achieves a 1.74\% increase in average OCR accuracy and a 2.50\% improvement in human-evaluated text correctness. 
The table illustrates that DiffCTE performs better than other approaches in terms of text generation accuracy and consistency. 
These results provide evidence of the effectiveness of DiffCTE in producing accurate and consistent text editing outcomes.

Additionally, we highlight the visual results of text editing in Figure~\ref{fig:vis}. 
As observed in Figure~\ref{fig:vis}, DiffCTE closely follows the provided text instructions, efficiently modifying the text content in the intended editing regions while maintaining style consistency.
For example, the ``DOWN'' generated by DiffCTE preserves the style consistency with the original image editing region, encompassing aspects such as font, color, and background texture.
In contrast, other methods occasionally generate inconsistent text, displaying noticeable variations in text style or background texture from the expected output. 
These findings suggest the advantage of DiffCTE in accomplishing both accurate text modification and consistent style preservation compared to other methods. 
The capacity of DiffCTE to generate visually coherent and stylistically consistent text editing demonstrates its potential and robustness in text editing tasks.

\subsection{Ablation Study}

To rigorously examine the effectiveness and potential future improvements of individual components within our proposed DiffCTE model, we carry out an extensive series of ablation studies. 
This analysis assesses the impact of the predicted editing region position and the influence of the style extractor.
Additionally, it evaluates complex character shape generation, particularly in Chinese, and explores the performance in style transfer tasks. 

\noindent\textbf{Effect of Predicted Editing Region Position}: In Table~\ref{tab:ablation_res}, by comparing the model with manually delineated editing region and the model employing text masks from predicting position of the editing region, we observe consistent text editing results.
This consistency is reflected in text generation accuracy and the naturalness of the generated text image, which validates the effectiveness of our model's editing region prediction and emphasizes the essential role of the text decoder in the VLM.

\noindent\textbf{Effect of Style Extractor}: Table~\ref{tab:ablation_res} and Figure~\ref{fig:without_sty} demonstrate that when the style condition is not incorporated and the latent vector derived from the text through the CLIP text encoder is used as the condition, the model's capacity to generate accurate text is reduced.
This leads to decreased OCR accuracy and Cor values. In contrast, incorporating the style extractor into the DiffCTE model results in a noticeable enhancement in performance.
This suggests that the style extractor contributes to providing prior knowledge of the style, ultimately improving text generation accuracy, consistency, and adaptability for various styles.

\noindent\textbf{Style Transfer Task}: The proposed DiffCTE model demonstrates effective performance in style transfer tasks, where a new image representing a specific style serves as a reference.
In contrast, DiffUTE is unable to perform style transfer from a reference image.
As shown in Figure~\ref{fig:ablation_task}, the model extracts style features from the reference image and modifies the original image's text while simultaneously altering the style.
This process illustrates the model's capability to perform style transfer while preserving the text's accuracy and stylistic consistency, indicating potential for further refinements in such tasks.

\noindent\textbf{Complex Character Generation}: As shown in Figure~\ref{fig:ablation_task}, our proposed DiffCTE model exhibits adaptability and proficiency in generating complex character shapes, particularly in languages such as Chinese.
Compared to DiffSTE, this highlights the model's capacity to handle diverse character complexities across different languages and reveals the potential for further advancements in this aspect.

\section{Conclusion}
In this paper, we present a novel method, DiffCTE, designed to tackle the challenge of text generation inconsistency in text editing tasks.
DiffCTE integrates the VLM and the diffusion model to generate high-quality, readable text images while preserving style consistency. 
The experimental results illustrate that DiffCTE attains remarkable fidelity and consistency in generating text. 
These findings underscore the efficacy and potential of DiffCTE in addressing the complexities associated with text editing tasks. 
The proposed method establishes a foundation for generating accurate and stylistically consistent text editing and serves as a catalyst for future research in this domain.
{
    \small
    \bibliographystyle{ieeenat_fullname}
    \bibliography{main}

@String(ICME = {Int. Conf. Multimedia and Expo})

@String(AAAI = {AAAI})

@String(ICME  =	{ICME})

@inproceedings{wu2019editing,
  title={Editing text in the wild},
  author={Wu, Liang and Zhang, Chengquan and Liu, Jiaming and Han, Junyu and Liu, Jingtuo and Ding, Errui and Bai, Xiang},
  booktitle={Proceedings of the 27th ACM international conference on multimedia},
  pages={1500--1508},
  year={2019}
}

@article{kang2021content,
  title={Content and style aware generation of text-line images for handwriting recognition},
  author={Kang, Lei and Riba, Pau and Rusinol, Marcal and Fornes, Alicia and Villegas, Mauricio},
  journal={IEEE Transactions on Pattern Analysis and Machine Intelligence},
  volume={44},
  number={12},
  pages={8846--8860},
  year={2021},
  publisher={IEEE}
}

@article{luo2021separating,
  title={Separating content from style using adversarial learning for recognizing text in the wild},
  author={Luo, Canjie and Lin, Qingxiang and Liu, Yuliang and Jin, Lianwen and Shen, Chunhua},
  journal={International Journal of Computer Vision},
  volume={129},
  pages={960--976},
  year={2021},
  publisher={Springer}
}

@article{yu2021mask,
  title={Mask-guided GAN for robust text editing in the scene},
  author={Yu, Boxi and Xu, Yong and Huang, Yan and Yang, Shuai and Liu, Jiaying},
  journal={Neurocomputing},
  volume={441},
  pages={192--201},
  year={2021},
  publisher={Elsevier}
}

@article{lee2021rewritenet,
  title={Rewritenet: Realistic scene text image generation via editing text in real-world image},
  author={Lee, Junyeop and Kim, Yoonsik and Kim, Seonghyeon and Yim, Moonbin and Shin, Seung and Lee, Gayoung and Park, Sungrae},
  journal={arXiv preprint arXiv:2107.11041},
  volume={1},
  year={2021}
}

@article{su2023scene,
  title={Scene Style Text Editing},
  author={Su, Tonghua and Yang, Fuxiang and Zhou, Xiang and Di, Donglin and Wang, Zhongjie and Li, Songze},
  journal={arXiv preprint arXiv:2304.10097},
  year={2023}
}

@inproceedings{roy2020stefann,
  title={STEFANN: scene text editor using font adaptive neural network},
  author={Roy, Prasun and Bhattacharya, Saumik and Ghosh, Subhankar and Pal, Umapada},
  booktitle={Proceedings of the IEEE/CVF Conference on Computer Vision and Pattern Recognition},
  pages={13228--13237},
  year={2020}
}

@inproceedings{qu2023exploring,
  title={Exploring stroke-level modifications for scene text editing},
  author={Qu, Yadong and Tan, Qingfeng and Xie, Hongtao and Xu, Jianjun and Wang, Yuxin and Zhang, Yongdong},
  booktitle={Proceedings of the AAAI Conference on Artificial Intelligence},
  volume={37},
  number={2},
  pages={2119--2127},
  year={2023}
}

@article{huang2022gentext,
  title={Gentext: Unsupervised artistic text generation via decoupled font and texture manipulation},
  author={Huang, Qirui and Fu, Bin and Zhang, Aozhong and Qiao, Yu},
  journal={arXiv preprint arXiv:2207.09649},
  year={2022}
}

@inproceedings{kong2022look,
  title={Look closer to supervise better: one-shot font generation via component-based discriminator},
  author={Kong, Yuxin and Luo, Canjie and Ma, Weihong and Zhu, Qiyuan and Zhu, Shenggao and Yuan, Nicholas and Jin, Lianwen},
  booktitle={Proceedings of the IEEE/CVF conference on computer vision and pattern recognition},
  pages={13482--13491},
  year={2022}
}

@inproceedings{shimoda2021rendering,
  title={De-rendering stylized texts},
  author={Shimoda, Wataru and Haraguchi, Daichi and Uchida, Seiichi and Yamaguchi, Kota},
  booktitle={Proceedings of the IEEE/CVF International Conference on Computer Vision},
  pages={1076--1085},
  year={2021}
}

@inproceedings{yang2020swaptext,
  title={Swaptext: Image based texts transfer in scenes},
  author={Yang, Qiangpeng and Huang, Jun and Lin, Wei},
  booktitle={Proceedings of the IEEE/CVF Conference on Computer Vision and Pattern Recognition},
  pages={14700--14709},
  year={2020}
}

@inproceedings{zhan2019spatial,
  title={Spatial fusion gan for image synthesis},
  author={Zhan, Fangneng and Zhu, Hongyuan and Lu, Shijian},
  booktitle={Proceedings of the IEEE/CVF conference on computer vision and pattern recognition},
  pages={3653--3662},
  year={2019}
}

@article{balaji2022ediffi,
  title={ediffi: Text-to-image diffusion models with an ensemble of expert denoisers},
  author={Balaji, Yogesh and Nah, Seungjun and Huang, Xun and Vahdat, Arash and Song, Jiaming and Kreis, Karsten and Aittala, Miika and Aila, Timo and Laine, Samuli and Catanzaro, Bryan and others},
  journal={arXiv preprint arXiv:2211.01324},
  year={2022}
}

@article{brack2023sega,
  title={Sega: Instructing diffusion using semantic dimensions},
  author={Brack, Manuel and Friedrich, Felix and Hintersdorf, Dominik and Struppek, Lukas and Schramowski, Patrick and Kersting, Kristian},
  journal={arXiv preprint arXiv:2301.12247},
  year={2023}
}

@article{ho2020denoising,
  title={Denoising diffusion probabilistic models},
  author={Ho, Jonathan and Jain, Ajay and Abbeel, Pieter},
  journal={Advances in neural information processing systems},
  volume={33},
  pages={6840--6851},
  year={2020}
}

@article{li2023guiding,
  title={Guiding text-to-image diffusion model towards grounded generation},
  author={Li, Ziyi and Zhou, Qinye and Zhang, Xiaoyun and Zhang, Ya and Wang, Yanfeng and Xie, Weidi},
  journal={arXiv preprint arXiv:2301.05221},
  year={2023}
}

@article{nichol2021glide,
  title={Glide: Towards photorealistic image generation and editing with text-guided diffusion models},
  author={Nichol, Alex and Dhariwal, Prafulla and Ramesh, Aditya and Shyam, Pranav and Mishkin, Pamela and McGrew, Bob and Sutskever, Ilya and Chen, Mark},
  journal={arXiv preprint arXiv:2112.10741},
  year={2021}
}

@inproceedings{nichol2021improved,
  title={Improved denoising diffusion probabilistic models},
  author={Nichol, Alexander Quinn and Dhariwal, Prafulla},
  booktitle={International Conference on Machine Learning},
  pages={8162--8171},
  year={2021},
  organization={PMLR}
}

@article{ryu2022pyramidal,
  title={Pyramidal denoising diffusion probabilistic models},
  author={Ryu, Dohoon and Ye, Jong Chul},
  journal={arXiv preprint arXiv:2208.01864},
  year={2022}
}

@article{wu2022unifying,
  title={Unifying Diffusion Models' Latent Space, with Applications to CycleDiffusion and Guidance},
  author={Wu, Chen Henry and De la Torre, Fernando},
  journal={arXiv preprint arXiv:2210.05559},
  year={2022}
}

@inproceedings{wu2023uncovering,
  title={Uncovering the disentanglement capability in text-to-image diffusion models},
  author={Wu, Qiucheng and Liu, Yujian and Zhao, Handong and Kale, Ajinkya and Bui, Trung and Yu, Tong and Lin, Zhe and Zhang, Yang and Chang, Shiyu},
  booktitle={Proceedings of the IEEE/CVF Conference on Computer Vision and Pattern Recognition},
  pages={1900--1910},
  year={2023}
}

@inproceedings{xie2023smartbrush,
  title={Smartbrush: Text and shape guided object inpainting with diffusion model},
  author={Xie, Shaoan and Zhang, Zhifei and Lin, Zhe and Hinz, Tobias and Zhang, Kun},
  booktitle={Proceedings of the IEEE/CVF Conference on Computer Vision and Pattern Recognition},
  pages={22428--22437},
  year={2023}
}

@inproceedings{brooks2023instructpix2pix,
  title={Instructpix2pix: Learning to follow image editing instructions},
  author={Brooks, Tim and Holynski, Aleksander and Efros, Alexei A},
  booktitle={Proceedings of the IEEE/CVF Conference on Computer Vision and Pattern Recognition},
  pages={18392--18402},
  year={2023}
}

@inproceedings{rombach2022high,
  title={High-resolution image synthesis with latent diffusion models},
  author={Rombach, Robin and Blattmann, Andreas and Lorenz, Dominik and Esser, Patrick and Ommer, Bj{\"o}rn},
  booktitle={Proceedings of the IEEE/CVF conference on computer vision and pattern recognition},
  pages={10684--10695},
  year={2022}
}

@inproceedings{DiffUTE,
      title={DiffUTE: Universal Text Editing Diffusion Model},
      author={Chen, Haoxing and Xu, Zhuoer and Gu, Zhangxuan and Lan, Jun and Zheng, Xing and Li, Yaohui and Meng, Changhua and Zhu, Huijia and Wang, Weiqiang},
      booktitle={Thirty-seventh Conference on Neural Information Processing Systems (NeurIPS)},
      year={2023}
}

@article{ji2023improving,
  title={Improving Diffusion Models for Scene Text Editing with Dual Encoders},
  author={Ji, Jiabao and Zhang, Guanhua and Wang, Zhaowen and Hou, Bairu and Zhang, Zhifei and Price, Brian and Chang, Shiyu},
  journal={arXiv preprint arXiv:2304.05568},
  year={2023}
}

@misc{li2021cdla,
  title={Cdla: A chinese document layout analysis (cdla) dataset},
  author={Li, Hang},
  year={2021}
}

@inproceedings{xu2022xfund,
  title={Xfund: A benchmark dataset for multilingual visually rich form understanding},
  author={Xu, Yiheng and Lv, Tengchao and Cui, Lei and Wang, Guoxin and Lu, Yijuan and Florencio, Dinei and Zhang, Cha and Wei, Furu},
  booktitle={Findings of the Association for Computational Linguistics: ACL 2022},
  pages={3214--3224},
  year={2022}
}

@inproceedings{park2019cord,
  title={CORD: a consolidated receipt dataset for post-OCR parsing},
  author={Park, Seunghyun and Shin, Seung and Lee, Bado and Lee, Junyeop and Surh, Jaeheung and Seo, Minjoon and Lee, Hwalsuk},
  booktitle={Workshop on Document Intelligence at NeurIPS 2019},
  year={2019}
}

@inproceedings{guo2019eaten,
  title={Eaten: Entity-aware attention for single shot visual text extraction},
  author={Guo, He and Qin, Xiameng and Liu, Jiaming and Han, Junyu and Liu, Jingtuo and Ding, Errui},
  booktitle={2019 International Conference on Document Analysis and Recognition (ICDAR)},
  pages={254--259},
  year={2019},
  organization={IEEE}
}

@inproceedings{zhong2019publaynet,
  title={Publaynet: largest dataset ever for document layout analysis},
  author={Zhong, Xu and Tang, Jianbin and Yepes, Antonio Jimeno},
  booktitle={2019 International Conference on Document Analysis and Recognition (ICDAR)},
  pages={1015--1022},
  year={2019},
  organization={IEEE}
}

@inproceedings{zhang2019icdar,
  title={Icdar 2019 robust reading challenge on reading chinese text on signboard},
  author={Zhang, Rui and Zhou, Yongsheng and Jiang, Qianyi and Song, Qi and Li, Nan and Zhou, Kai and Wang, Lei and Wang, Dong and Liao, Minghui and Yang, Mingkun and others},
  booktitle={2019 international conference on document analysis and recognition (ICDAR)},
  pages={1577--1581},
  year={2019},
  organization={IEEE}
}

@inproceedings{nayef2019icdar2019,
  title={ICDAR2019 robust reading challenge on multi-lingual scene text detection and recognition—RRC-MLT-2019},
  author={Nayef, Nibal and Patel, Yash and Busta, Michal and Chowdhury, Pinaki Nath and Karatzas, Dimosthenis and Khlif, Wafa and Matas, Jiri and Pal, Umapada and Burie, Jean-Christophe and Liu, Cheng-lin and others},
  booktitle={2019 International conference on document analysis and recognition (ICDAR)},
  pages={1582--1587},
  year={2019},
  organization={IEEE}
}

@inproceedings{karatzas2015icdar,
  title={ICDAR 2015 competition on robust reading},
  author={Karatzas, Dimosthenis and Gomez-Bigorda, Lluis and Nicolaou, Anguelos and Ghosh, Suman and Bagdanov, Andrew and Iwamura, Masakazu and Matas, Jiri and Neumann, Lukas and Chandrasekhar, Vijay Ramaseshan and Lu, Shijian and others},
  booktitle={2015 13th international conference on document analysis and recognition (ICDAR)},
  pages={1156--1160},
  year={2015},
  organization={IEEE}
}

@inproceedings{huang2019icdar2019,
  title={Icdar2019 competition on scanned receipt ocr and information extraction},
  author={Huang, Zheng and Chen, Kai and He, Jianhua and Bai, Xiang and Karatzas, Dimosthenis and Lu, Shijian and Jawahar, CV},
  booktitle={2019 International Conference on Document Analysis and Recognition (ICDAR)},
  pages={1516--1520},
  year={2019},
  organization={IEEE}
}

@inproceedings{chng2019icdar2019,
  title={Icdar2019 robust reading challenge on arbitrary-shaped text-rrc-art},
  author={Chng, Chee Kheng and Liu, Yuliang and Sun, Yipeng and Ng, Chun Chet and Luo, Canjie and Ni, Zihan and Fang, ChuanMing and Zhang, Shuaitao and Han, Junyu and Ding, Errui and others},
  booktitle={2019 International Conference on Document Analysis and Recognition (ICDAR)},
  pages={1571--1576},
  year={2019},
  organization={IEEE}
}

@inproceedings{gomez2017icdar2017,
  title={Icdar2017 robust reading challenge on coco-text},
  author={Gomez, Raul and Shi, Baoguang and Gomez, Lluis and Numann, Lukas and Veit, Andreas and Matas, Jiri and Belongie, Serge and Karatzas, Dimosthenis},
  booktitle={2017 14th IAPR International Conference on Document Analysis and Recognition (ICDAR)},
  volume={1},
  pages={1435--1443},
  year={2017},
  organization={IEEE}
}

@inproceedings{singh2021textocr,
  title={Textocr: Towards large-scale end-to-end reasoning for arbitrary-shaped scene text},
  author={Singh, Amanpreet and Pang, Guan and Toh, Mandy and Huang, Jing and Galuba, Wojciech and Hassner, Tal},
  booktitle={Proceedings of the IEEE/CVF conference on computer vision and pattern recognition},
  pages={8802--8812},
  year={2021}
}

@article{cao2023mobile,
  title={Mobile augmented reality: User interfaces, frameworks, and intelligence},
  author={Cao, Jacky and Lam, Kit-Yung and Lee, Lik-Hang and Liu, Xiaoli and Hui, Pan and Su, Xiang},
  journal={ACM Computing Surveys},
  volume={55},
  number={9},
  pages={1--36},
  year={2023},
  publisher={ACM New York, NY}
}

@inproceedings{kwon2022clipstyler,
  title={Clipstyler: Image style transfer with a single text condition},
  author={Kwon, Gihyun and Ye, Jong Chul},
  booktitle={Proceedings of the IEEE/CVF Conference on Computer Vision and Pattern Recognition},
  pages={18062--18071},
  year={2022}
}

@article{shi2022aprnet,
  title={APRNet: Attention-based Pixel-wise Rendering Network for Photo-Realistic Text Image Generation},
  author={Shi, Yangming and Ding, Haisong and Chen, Kai and Huo, Qiang},
  journal={arXiv preprint arXiv:2203.07705},
  year={2022}
}

@article{liu2020erasenet,
  title={EraseNet: End-to-end text removal in the wild},
  author={Liu, Chongyu and Liu, Yuliang and Jin, Lianwen and Zhang, Shuaitao and Luo, Canjie and Wang, Yongpan},
  journal={IEEE Transactions on Image Processing},
  volume={29},
  pages={8760--8775},
  year={2020},
  publisher={IEEE}
}

@inproceedings{wang2023letter,
  title={Letter Embedding Guidance Diffusion Model for Scene Text Editing},
  author={Wang, Changshuo and Wu, Lei and Chen, Xu and Li, Xiang and Meng, Lei and Meng, Xiangxu},
  booktitle={2023 IEEE International Conference on Multimedia and Expo (ICME)},
  pages={588--593},
  year={2023},
  organization={IEEE}
}

@article{chen2023textdiffuser,
  title={TextDiffuser: Diffusion Models as Text Painters},
  author={Chen, Jingye and Huang, Yupan and Lv, Tengchao and Cui, Lei and Chen, Qifeng and Wei, Furu},
  journal={arXiv preprint arXiv:2305.10855},
  year={2023}
}

@article{hertz2022prompt,
  title={Prompt-to-prompt image editing with cross attention control},
  author={Hertz, Amir and Mokady, Ron and Tenenbaum, Jay and Aberman, Kfir and Pritch, Yael and Cohen-Or, Daniel},
  journal={arXiv preprint arXiv:2208.01626},
  year={2022}
}

@article{tuo2023anytext,
  title={AnyText: Multilingual Visual Text Generation And Editing},
  author={Tuo, Yuxiang and Xiang, Wangmeng and He, Jun-Yan and Geng, Yifeng and Xie, Xuansong},
  journal={arXiv preprint arXiv:2311.03054},
  year={2023}
}

@article{ma2023glyphdraw,
  title={GlyphDraw: Learning to Draw Chinese Characters in Image Synthesis Models Coherently},
  author={Ma, Jian and Zhao, Mingjun and Chen, Chen and Wang, Ruichen and Niu, Di and Lu, Haonan and Lin, Xiaodong},
  journal={arXiv preprint arXiv:2303.17870},
  year={2023}
}

@article{li2023blip,
  title={Blip-diffusion: Pre-trained subject representation for controllable text-to-image generation and editing},
  author={Li, Dongxu and Li, Junnan and Hoi, Steven CH},
  journal={arXiv preprint arXiv:2305.14720},
  year={2023}
}

@inproceedings{kim2022ocr,
  title={Ocr-free document understanding transformer},
  author={Kim, Geewook and Hong, Teakgyu and Yim, Moonbin and Nam, JeongYeon and Park, Jinyoung and Yim, Jinyeong and Hwang, Wonseok and Yun, Sangdoo and Han, Dongyoon and Park, Seunghyun},
  booktitle={European Conference on Computer Vision},
  pages={498--517},
  year={2022},
  organization={Springer}
}
}

\clearpage
\setcounter{page}{1}
\maketitlesupplementary

\section{More Experimental Results}
\label{sec:rationale}
\begin{figure}
\centering
    \includegraphics[width=0.50\textwidth,trim=310 70 300 70, clip]{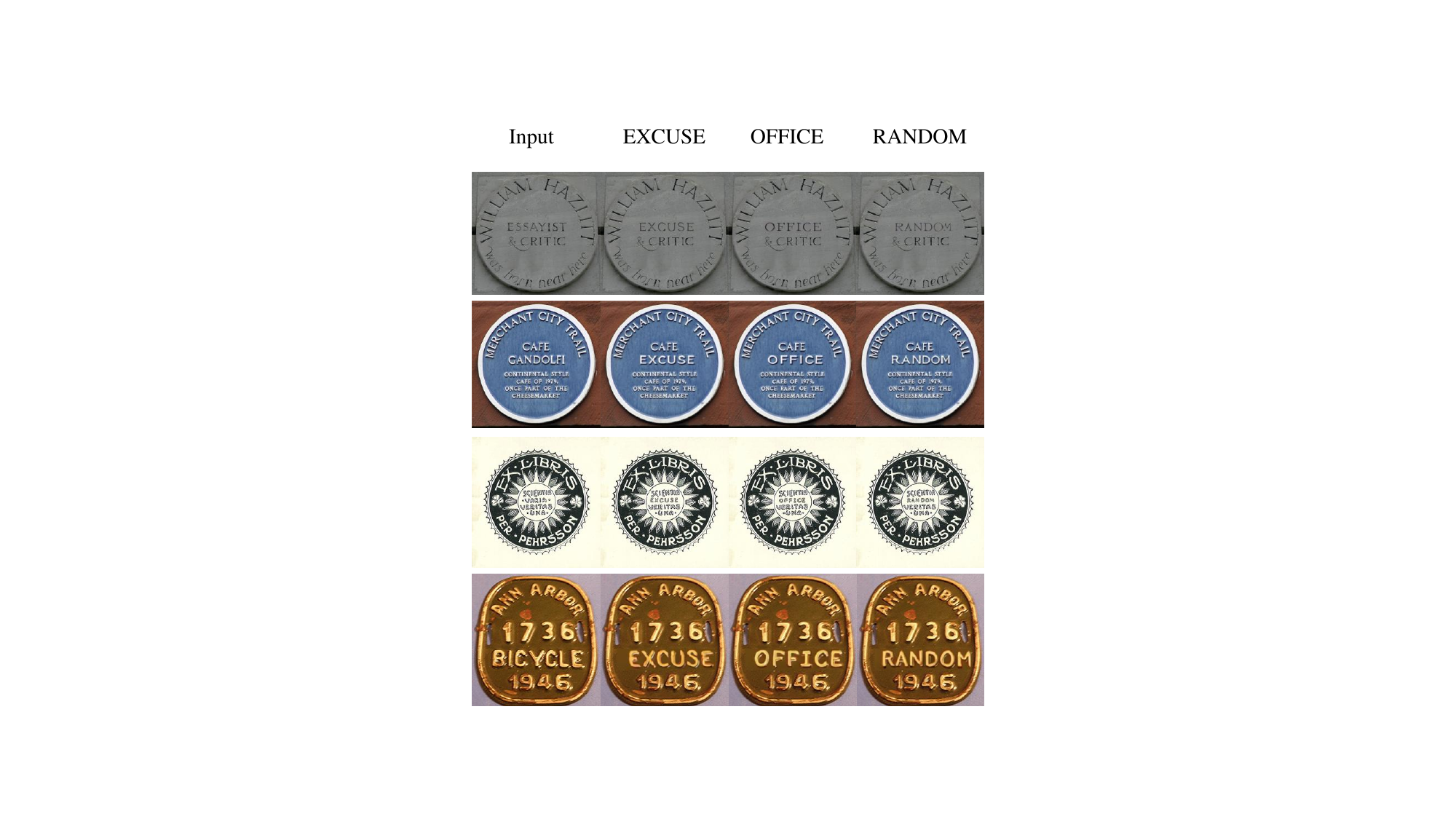}
    \caption{Realistic style consistency word substitution using our proposed DiffCTE. Each row represents a different style, and each column represents different edited text. Our DiffCTE demonstrates the ability to preserve style consistency in the generated text image.}
    \label{fig:sup_styleconsistency}
\end{figure}

\begin{figure}
  \includegraphics[width=0.50\textwidth,trim=290 78 280 80, clip]{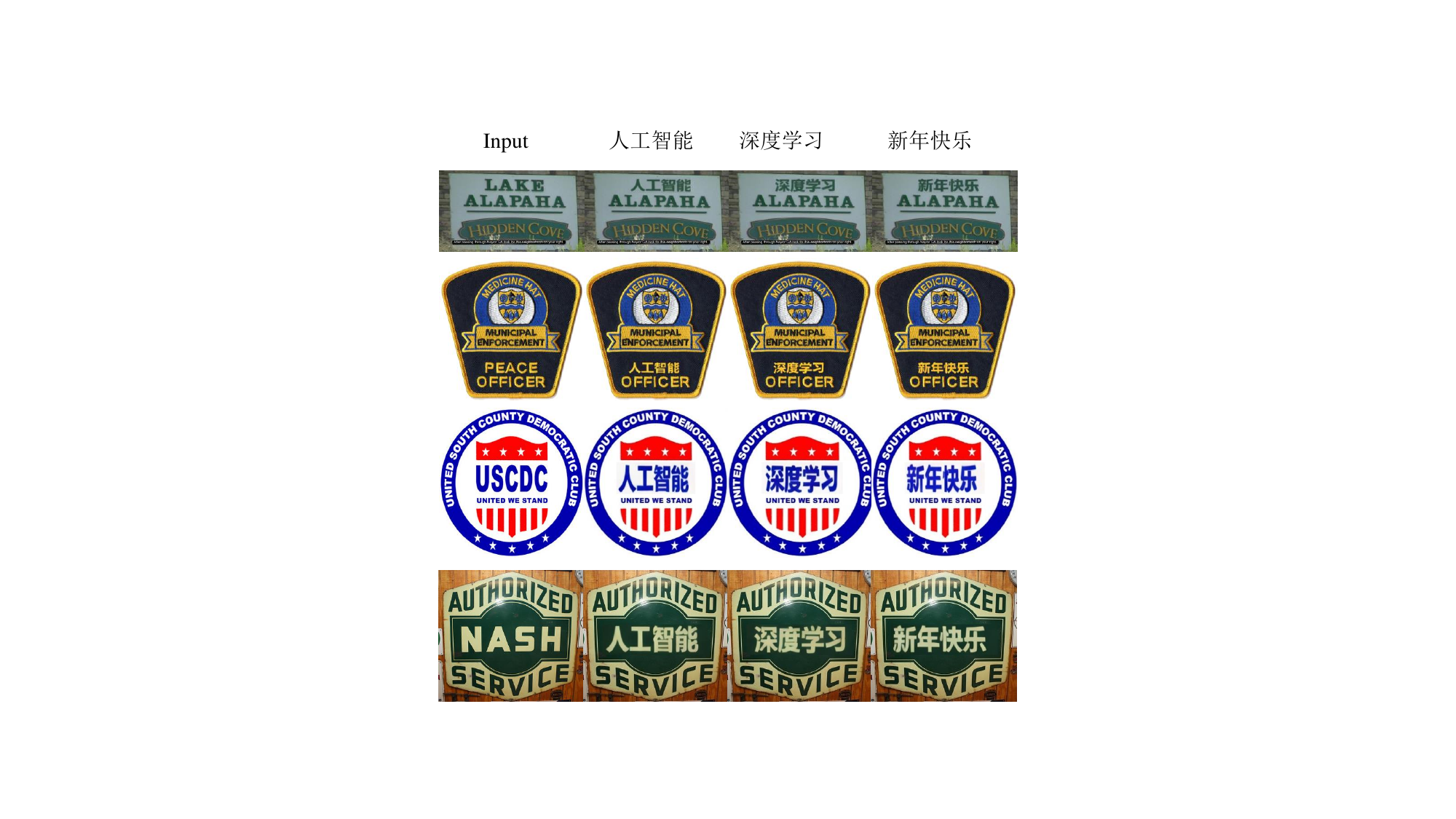}
  \caption{Complex character generation using our proposed DiffCTE. The model adeptly transfers style from English characters to Chinese characters, demonstrating its versatility. In the second to fourth columns, the Chinese characters correspond to the phrases ``Artificial Intelligence", ``Deep Learning," and ``Happy New Year", respectively.}
  \label{fig:sup_complexchar}
\end{figure}

%
For an in-depth examination of the DiffCTE model, additional edited images are presented in Figure~\ref{fig:sup_styleconsistency}, Figure~\ref{fig:sup_complexchar}, and Figure~\ref{fig:sup_badcase}. These figures depict realistic style consistency in word substitution, complex character generation, and model limitations, respectively.

Figures~\ref{fig:sup_styleconsistency} and \ref{fig:sup_complexchar} illustrate the effectiveness of the DiffCTE model in generating diverse and visually consistent characters, emphasizing its ability to capture distinct word styles. Conversely, Figure~\ref{fig:sup_badcase} underlines some challenges encountered by DiffCTE, such as generating characters with high visual similarity or handling cases involving repetitive texts.

\subsection{Style Consistency Word Substitution Analysis}
Figure~\ref{fig:sup_styleconsistency} showcases four rows of images, each representing different styles: carved, raised, hollow, and metallic characters, which are prevalent in everyday life.

Upon modifying the text of the target region in the input image to ``EXCUSE", ``OFFICE", and ``RANDOM", the output images demonstrate maintained style consistency. Importantly, the DiffCTE model exhibits a capacity for learning subtle word styles from the input image.

For instance, the first row image displays the word ``ESSAYIST" slightly blurred due to erosion, which DiffCTE consistently preserves. 
The second row images exhibit a faint shading effect, creating a raised style. 
The third row images feature hollow, distorted characters, indicating the model's ability to learn complex text styles from various inputs. 
Lastly, the fourth row presents metallic characters in different forms, depending on light direction and intensity, as well as the character's protrusion.

\subsection{Complex Character Generation Assessment}
In Figure~\ref{fig:sup_complexchar}, four rows of images are presented. 
Although the input images contain only English words, DiffCTE effectively learns the style from the English text and successfully transfers it to Chinese characters.

The resulting Chinese characters display consistent stroke border width and color with the original English characters, suggesting DiffCTE's adaptability in learning stroke thickness and color attributes. 
Furthermore, the Chinese characters obtain additional features influenced by the input images. 
For example, characters in the second row demonstrate a gradient shading effect. 
Characters within the same column reveal subtle variations imbued by the input image style.

\subsection{Failure Generation Analysis}
\begin{figure}
  \includegraphics[width=0.50\textwidth,trim=245 110 240 80, clip]{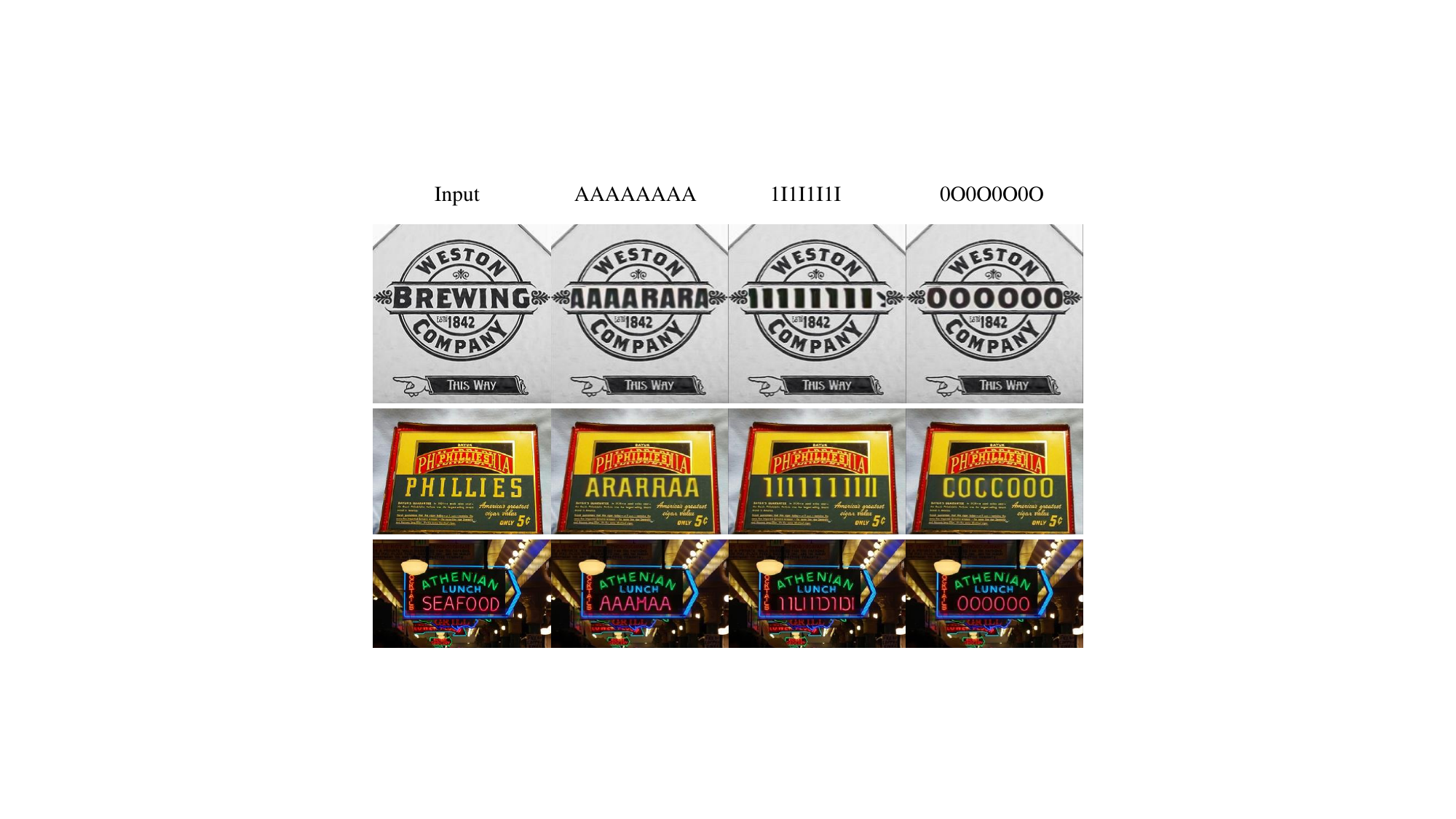}
  \caption{Failure cases generated using our proposed DiffCTE. DiffCTE exhibits limitations in handling cases involving repetitive texts or generating characters with high visual similarity.}
  \label{fig:sup_badcase}
  \vspace{3.0cm}
\end{figure}

Figure~\ref{fig:sup_badcase} highlights instances where DiffCTE generates suboptimal results. From the second column, it is evident that the English letter ``A'' is occasionally generated as ``R''. Similarly, the digit ``1'' and the English letter ``I'' exhibit visual similarity, leading to potential confusion during the generation process. Likewise, the digit ``0'' and the English letter ``O'' display notable visual similarities, which can introduce ambiguity in the output.

By examining these scenarios, we can identify potential directions for future research, such as enhancing the model's ability to handle visually similar characters and repetitive texts. The provided analysis offers a comprehensive understanding of the DiffCTE model's capabilities as well as its limitations, paving the way for opportunities to refine and extend the model's performance in various applications.

\end{document}